\documentclass[a4paper,twoside]{article}

\usepackage{epsfig}
\usepackage{subcaption}
\usepackage{calc}
\usepackage{amssymb}
\usepackage{amstext}
\usepackage{amsmath}
\usepackage{amsthm}
\usepackage{multicol}
\usepackage{pslatex}
\usepackage{apalike}
\usepackage{SCITEPRESS}     
\usepackage{multirow}

\begin{document}

\title{VMRFANet: View-Specific Multi-Receptive Field Attention Network for Person Re-identification}
\author{\authorname{Honglong Cai, Yuedong Fang, Zhiguan Wang, Tingchun Yeh, Jinxing Cheng}
\affiliation{Suning Commerce R\&D Center USA}
\email{\{honglong.cai, yuedong.fang, doris.wang, tingchun.yeh, jim.cheng\}@ussuning.com}
}


\abstract{Person re-identification (re-ID) aims to retrieve the same person across different cameras. In practice, it still
remains a challenging task due to background clutter, variations on body poses and view conditions, inaccurate
bounding box detection, etc. To tackle these issues, in this paper, we propose a novel multi-receptive field
attention (MRFA) module that utilizes filters of various sizes to help network focusing on informative pixels.
Besides, we present a view-specific mechanism that guides attention module to handle the variation of view
conditions. Moreover, we introduce a Gaussian horizontal random cropping/padding method which further
improves the robustness of our proposed network. Comprehensive experiments demonstrate the effectiveness of each component. Our method achieves 95.5\% / 88.1\%
in rank-1 / mAP on Market-1501, 
88.9\% / 80.0\% on DukeMTMC-reID, 81.1\% / 78.8\% on CUHK03 labeled dataset and 78.9\% / 75.3\% on CUHK03 detected dataset,  outperforming current state-of-the-art methods.}

\onecolumn \maketitle \normalsize \setcounter{footnote}{0} \vfill

\section{\uppercase{Introduction}}
\label{sec:introduction}

Image-based person re-identification (re-ID) aims to search people from a large number of bounding boxes that have been detected across different cameras. Although extensive amounts of efforts and progress have been made in the past few years, person re-ID remains a challenging task in computer vision. The obstacles mainly come from the low resolution of images, background clutter, variations of person poses, \textit{etc}.

 Nowadays, the extracted deep features of pedestrian bounding boxes through a convolutional neural network(CNN) is demonstrated to be more discriminative and robust. However, most of the existing methods only learn global features from whole human body images such that some local discriminative information of specific parts may be ignored. To address this issue, some recent works \cite{pcb,mgn,alignedreid} archived state-of-the-art performance by dividing the extracted human image feature map into horizontal stripes and aggregating local representations from these fixed parts. Nevertheless, drawbacks of these part-based models are still obvious: 1) Feature units within each local feature map are treated equally by applying global average/maximum pooling to get refined feature representation. Thus the resulting models cannot focus more on discriminative local regions. And 2) Pre-defined feature map partition strategies are likely to suffer from misalignment issues. For example, the performance of methods adopting equal partition strategies (e.g. \cite{pcb}) heavily depends on the quality and robustness of pedestrian bounding box detection, which itself is a challenging task. Other strategies such as partition based on human pose (e.g. \cite{YANG2019143}) often introduce side models trained on different datasets. In that case, domain bias may come into play.
 
 Moreover, to our best knowledge, none of these methods have made efforts to manage view-specific bias. That is, the variation of view conditions from different cameras can be dramatic. Thus the extracted features are likely to be biased in a way that intra-class features of images from different views will be pushed apart, and inter-class ones from the same view will be pulled closer. To better handle these problems, adopting an attention mechanism is an intuitive and effective choice. As human vision only focuses on selective parts instead of processing the whole field of view at once, attention mechanism aims to detect informative pixels within an image. It can help to extract features that better represent the regions of interest while suppressing the non-target regions. Meanwhile, it can be trained along with the feature extractor in an end-to-end manner. 
 
 In this work, we explore the application of attention mechanisms on the person re-identification problem. Particularly, the contributions of this paper can be summarized as follow: 
\begin{itemize}
    \item We investigate the idea of combining spatial- and channel-wise attention in a single module with various sized receptive filters, and then mount the module to a popular strip-based re-ID baseline \cite{pcb} in a parallel way. We believe this is a more general form of attention module comparing to the ones in many existing structures that try to learn spatial- and channel-wise attention separately.
    \item We explore the potential of using attention module to inject prior information into feature extractor. To be specific, we utilize the camera ID tag to guide our attention module learning a view specific feature mask that further improves the re-ID performance.
    \item We propose a novel horizontal data augmentation technique against the misalignment risk, which is a well-known shortcoming of strip-based models.
\end{itemize}

\section{\uppercase{Related Work}}

\noindent\textbf{Strip-based models:}
Recently, strip-based models have been proven to be effective in person re-ID. Part-based Convolutional Baseline (PCB) \cite{pcb} equally slices the final feature map into horizontal strips. After refining part pooling, the extracted local features are jointly trained with classification losses and have been concatenated as the final feature. Lately, \cite{mgn} proposed a multi-branch network to combine global and partial features at different granularities. With the combination of classification and triplet losses, it pushed the re-ID performances to a new level compared with previous state-of-the-art methods. Due to the effectiveness and simplicity, we adopted a modified version of PCB structure as the baseline in this work.

\noindent\textbf{Attention mechanism in Re-ID:}
Another challenge in person re-ID is imperfect bounding-box detection. To address this issue, the attention mechanism is a natural choice for aiding the network to learn where to ``look'' at. There are a few attempts in the literature that apply attention mechanisms for solving re-ID task \cite{Maskguided,YANG2019143,hacnn,mlfn}. For example, \cite{Maskguided} utilized body masks to guide the training of attention module. \cite{YANG2019143} proposed an end-to-end trainable framework composed of local and fusion attention modules that can incorporate image partition using human key-points estimation. Our proposed MRFA module is designed to address the imperfect detection issue mentioned above. Meanwhile, unlike \cite{hacnn} and a few other existing attention-based methods, MRFA tries to preserve the cross-correlation between spatial- and channel-wise attention.


\noindent\textbf{Metric learning:}
 Metric learning projects images to a vector space with fixed dimensions and defines a metric to compute distances between embedded features. one direction is to study the distance function explicitly. A representative and illuminating example is \cite{yu18}: to tackle the unsupervised re-ID problem, they proposed a deep framework consisting of a CNN feature extractor and an asymmetric metric layer such that the feature from extractor will be transformed specifically according to the view to form the final feature in Euclidean space. Like many other re-ID methods, we also incorporate the triplet loss in this work to enhance the feature representability. Besides, we also investigate the usage of attention module acting like the asymmetric metric layer to learn a view-specific attention map.


\section{\uppercase{The Proposed Method}}
\begin{figure*}[ht]
    \centering
    \includegraphics[width=\textwidth]{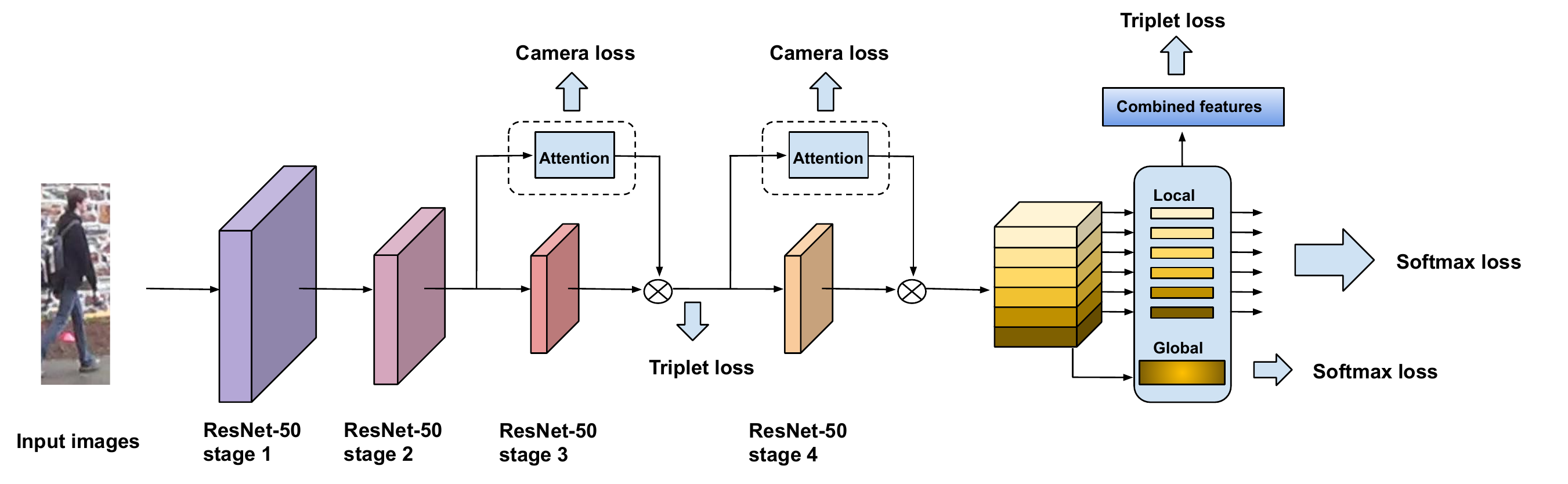}
    \caption{The structure of the proposed network (VMRFANet). Two attention modules are mounted to the third and fourth stages of ResNet50 backbone. Six local features are extracted from the last feature map together with a global feature. All seven features are concatenated and normalized to form a final descriptor of a pedestrian bounding box.}
    \label{fig:network}
\end{figure*}


In this section, we propose a novel attention module as well as a framework to train view specific feature enhancement/attenuation using the attention mechanism. A data augmentation method to improve the robustness of strip-based models has also been presented.  

\subsection{Overall Architecture}\label{sec:architecture}

The overall architecture of our proposed model is shown in Figure~\ref{fig:network}.

\noindent\textbf{Baseline network:} In this paper, we employ ResNet50 \cite{he15} as a backbone network with some modifications following \cite{pcb}: the last average pooling and fully connected layers have been removed as well as the down-sampling operation at the first layer of stage 5. We denote the dimension of the final feature map as $C\times H\times W$, where $C$ is the encoded channel dimension, and $H,W$ are the height and width respectively. A feature extractor has been applied to the final feature map to get a 512-dimensional global feature vector. Just like PCB, we further divide the final feature map into 6 horizontal strips such that each strip is of dimension $C\times (H/6) \times W$. Then each strip is fed to a feature extractor, so we end up getting 6 local feature vectors in total with dimension 256 each. Afterward, each feature is input to a fully-connected (FC) layer and the following Softmax function to classify the identities of the input images. Finally, all 7 feature vectors (6 local and 1 global) are concatenated to form a 2048-dimensional feature vector for inference.

\noindent\textbf{Other components:}
Two Multi-Receptive Field Attention (MRFA) modules, which will be described later in detail in Section~\ref{sec:attention_module}, are added to the baseline network. The first attention module takes the feature map after stage 2 block as an input. Its output mask $m_1\in (0, 1)^{1024 \times 24 \times 8}$ is then applied to the feature map after stage 3 block by an element-wise multiplication. The second attention module is mounted to stage 4 block similarly. Additionally, a feature extractor is connected to each attention module to extract a 512-dimensional feature for camera view classification, which will be explained in detail in Section~\ref{sec:view_specific}. 


\begin{figure}[ht]
\centering
    \includegraphics[width=0.46\textwidth]{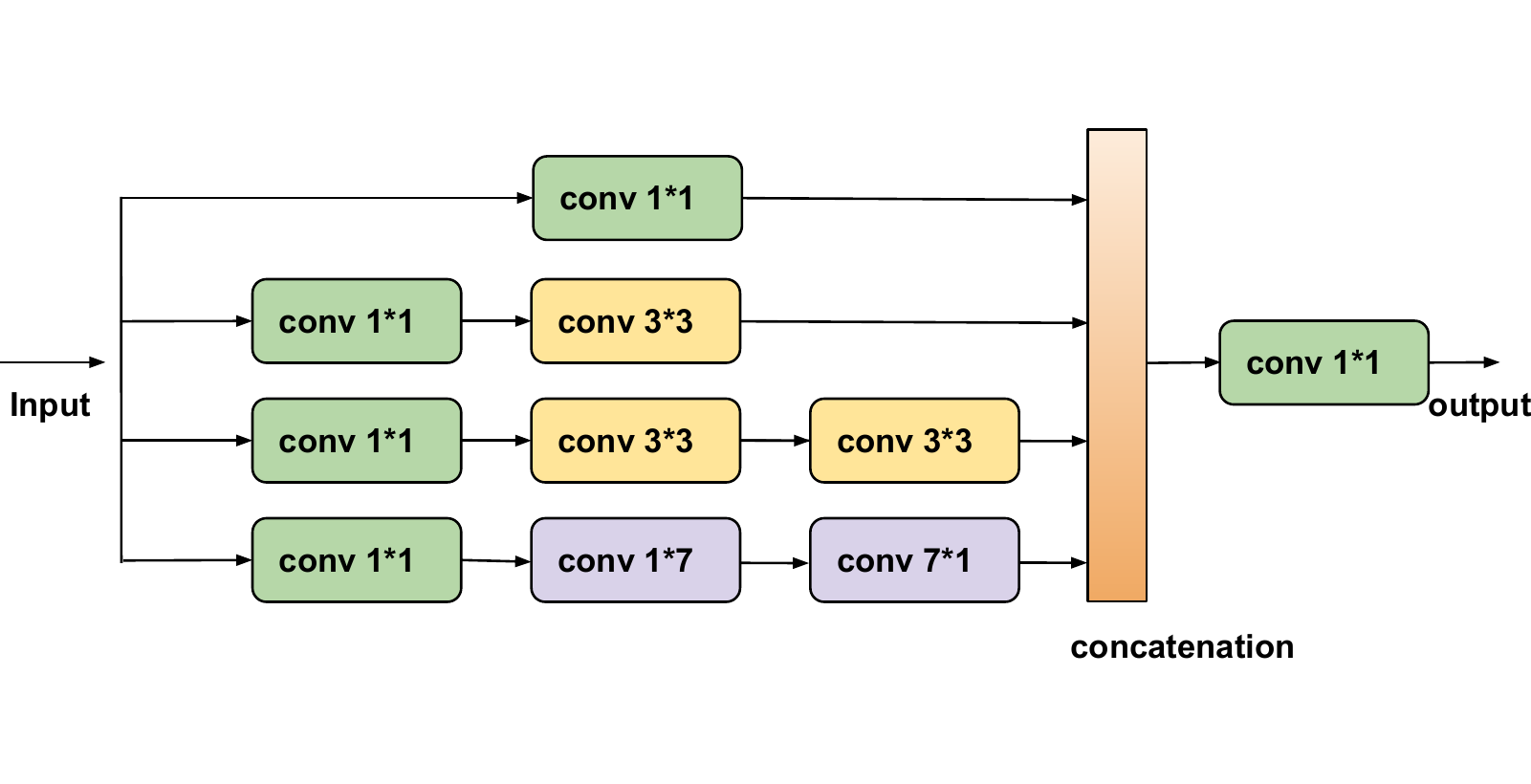}
    \caption{The detailed structure of a Multi-Receptive Field Attention (MRFA) module.}
    \label{fig:MRFA}
\end{figure}

\begin{figure}
    \centering
    \begin{subfigure}{0.07\textwidth}
        \includegraphics[width=\textwidth,height=0.10\textheight]{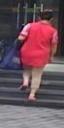}
        \caption{}
    \end{subfigure}
    \begin{subfigure}{0.07\textwidth}
        \includegraphics[width=\textwidth,height=0.10\textheight]{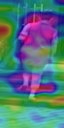}
        \caption{}
    \end{subfigure}
    \begin{subfigure}{0.07\textwidth}
        \includegraphics[width=\textwidth,height=0.10\textheight]{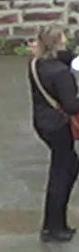}
        \caption{}
    \end{subfigure}
    \begin{subfigure}{0.07\textwidth}
        \includegraphics[width=\textwidth,height=0.10\textheight]{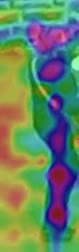}
        \caption{}
    \end{subfigure}
    \begin{subfigure}{0.07\textwidth}
        \includegraphics[width=\textwidth,height=0.10\textheight]{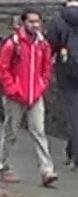}
        \caption{}
    \end{subfigure}
    \begin{subfigure}{0.07\textwidth}
        \includegraphics[width=\textwidth,height=0.10\textheight]{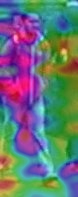}
        \caption{}
    \end{subfigure}
    \caption{Attention map of our MRFA module. (a) (c) (e) show the original images and (b) (d) (f) illustrate the corresponding attention maps. Attention maps show that our attention mechanism can focus on the person and filter out the background noise.}
    \label{fig:attentionmap}
\end{figure}

\subsection{Multi-Receptive Field Attention Module (MRFA)}\label{sec:attention_module}

To design the attention module, we use an Inception-like \cite{labelsmooth} architecture. That is, we design a shallow network with only up to four convolutional layers. Meanwhile, various filter sizes ($1 \times 1$, $3 \times 3$, $5 \times 5$, $7 \times 7$) have been adopted. And following \cite{labelsmooth}, we further reduce the number of parameters by factorizing convolutions with large filters of sizes $5 \times 5$ and $7 \times 7$ into two smaller $3 \times 3$ filters, and two asymmetric filters of sizes $1 \times 7$ and $7 \times 1$, respectively. The structure of MRFA is shown in Figure~\ref{fig:MRFA}. Our proposed attention structure can combine different reception field information and learn a different level of knowledge to make a decision which region we should pay more attention to. Figure~\ref{fig:attentionmap} shows that our attention mechanism can focus on the person's body and filter out background noise.

The input feature of channel dimension $C$ is first convolved by four $1 \times 1$ filters to be divided into four sub-features with channel dimension $C/4$ each. Then each sub-feature (except the one in the $1 \times 1$ filter branch) goes through filters of different sizes. For each filter, appropriate padding is applied to ensure the invariant of spatial dimensions.  Finally, all four sub-features will be concatenated to form a feature of channel dimension $C$, followed by a $1 \times 1$ convolution to be up-sampled to channel dimension $2C$ to match the channel size of feature from backbone network. A $tanh + 1$ function will be applied elemental-wise on the output attention map to normalize it to the range of $(0, 2)$. Note that due to spatial down-sampling at the beginning of stage 3 block, we need to apply average pooling after each $1 \times 1$ filter to ensure the matching of spatial dimensions between attention mask and feature map from backbone network.

\subsection{View Specific Learning Through Attention Mechanism}\label{sec:view_specific}
\begin{figure}
\centering
\begin{subfigure}[t]{0.08\textwidth}
    \makebox[0pt][r]{\makebox[30pt]{\raisebox{40pt}{\textbf{(a)}}}}%
    \includegraphics[width=\textwidth,height=0.10\textheight]{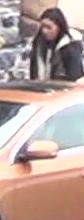}
    \makebox[0pt][r]{\makebox[30pt]{\raisebox{40pt}{\textbf{(b)}}}}%
    \includegraphics[width=\textwidth,height=0.10\textheight]{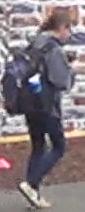}
    \makebox[0pt][r]{\makebox[30pt]{\raisebox{40pt}{\textbf{(c)}}}}%
    \includegraphics[width=\textwidth,height=0.10\textheight]{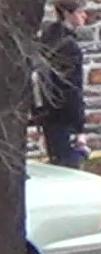}
\end{subfigure}
\hspace{0.1em}
\begin{subfigure}[t]{0.08\textwidth}
    \includegraphics[width=\textwidth,height=0.10\textheight]{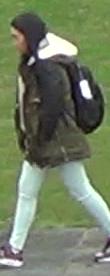}
    \includegraphics[width=\textwidth,height=0.10\textheight]{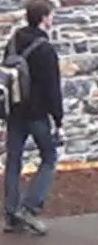}
    \includegraphics[width=\textwidth,height=0.10\textheight]{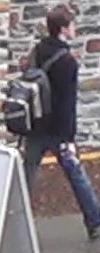}
\end{subfigure}
\hspace{0.1em}
\begin{subfigure}[t]{0.08\textwidth}
    \includegraphics[width=\textwidth,height=0.10\textheight]{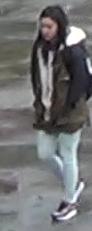}
    \includegraphics[width=\textwidth,height=0.10\textheight]{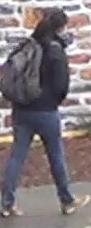}
    \includegraphics[width=\textwidth,height=0.10\textheight]{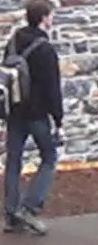}
\end{subfigure}
\caption{Example images from DukeMTMC-reID. (a) show bounding boxes of the same person captured by three different cameras. The included backgrounds and the view conditions various dramatically. (b) correspond to three different identities captured by a single camera such that they appear to be visually similar. (c) indicate the case of within-view inconsistency, i.e., the same person was captured by the same camera with different occlusions.}
\label{fig:view_specific}
\end{figure}

Our goal is to match people across different camera views distributed at different locations. The variation of cross-view person appearances can be dramatic due to various viewpoints, illumination conditions, and occlusion. As we can see, the same person looks different under different cameras and different persons look similar under same camera in Figure~\ref{fig:view_specific}



To tackle this issue, we thought it's effective to utilize the view-specific transformation. To make our network be aware of different camera views, we force our model to ``know'' which view the input bounding box belongs to. As a result, this task is converted to a camera ID (view) classification problem. However, in person re-ID task, the goal is to learn a camera-invariant feature which contradicts with camera ID (view) classification. To utilize the camera-specific information without affecting learning a camera-invariant final feature, we found it is natural to incorporate the view-specific transformation into our attention mechanism instead of adding on the backbone network. By adding camera ID (view) classification on the attention mechanism, we make it be aware of the view-specific information and could focus on the right place without affecting the camera-invariant features extracted from the backbone network.

This distance can be written as:
\begin{equation}
    d_l(\{\pmb{x}_i,v_i\},\{\pmb{x}_j, v_j\})=\lVert \pmb{U}^{\rm T}_{v_i}\pmb{x}_i-\pmb{U}^{\rm T}_{v_j}\pmb{x}_j\rVert_2
\end{equation}
where $\pmb{x}_i$ is the extracted feature of $i$-th bounding box, $v_i$ denotes the corresponding index of camera view, and $\pmb{U}_{v_i}$ is the view-specific transformation.


By connecting a simple feature extractor to each attention module, we denote the extracted attention feature $k$($k=1,2$) as $\pmb{a}_k$. We further add a fully connected layer to each feature extractor, the softmax loss is formulated as:
\begin{equation} \label{cam_id_loss}
    L^{\rm {softmax}}_{\rm {camera}} = -\frac{1}{N}\sum^{N}_{i=1}\sum^{2}_{k=1}{\log\frac{\exp(\pmb {W}^T_{v_i}\pmb{a}^i_k)}{\sum^{N_v}_{j=1}\exp(\pmb{W}^T_{j}\pmb{a}^i_k)}}
\end{equation}
where $\pmb{W}_j$ corresponds to the weight vector for camera ID $j$, with the size of mini-batch N and the number of cameras in the dataset $N_v$.

There remains one issue that needs to be dealt with carefully: the within-view inconsistency (see row (c) in Figure~\ref{fig:view_specific}), which arises when bounding boxes are detected at different locations within frames captured by the same camera. In that case, the view conditions can be distinct since different parts of the background will be included. To address this issue, we adopt a label smoothing \cite{labelsmooth} strategy on the softmax loss in Equation~\ref{cam_id_loss}: for a training example with ground-truth label $v_i$, we modify the label distribution $q(j)$ as:
\begin{equation}
    q^{\prime}(j)=(1-\epsilon)\delta_{j,v_i}+\frac{\epsilon}{N_v}
\end{equation}
Here $\delta_{j,v_i}$ is the Kronecker delta function and $(1-\epsilon)$ controls the level of confidence of the view classification. Thus the final loss function for view-specific learning can be written as:
\begin{equation}\label{cam_id_loss_smooth}
    L_{\rm{camera}} = -\frac{1}{N}\sum^{N}_{i=1}\sum^{2}_{k=1}\sum^{N_v}_{j=1}{\log p(j)q^{\prime}(j)}
\end{equation}
Where $p(j)$ is the predicted probability which is calculated by applying the softmax function on the output vector of the fully connected layer. 

\subsection{Combined loss}

Person re-identification is essentially a zero-shot learning task that identities in the training set will not overlap with those in the test set. But in order to let the network learn discriminate features, we can still formulate it as a multi-label classification problem by applying a softmax cross-entropy loss:
\begin{equation} \label{id_loss}
    L_{\rm ID} = -\frac{1}{N}\sum^{7}_{k=1}\sum^{N}_{i=1}{\log\frac{\exp(\pmb{W}^T_{y_j, k}\pmb{x}^i_k)}{\sum^{C}_{j=1}\exp(\pmb{W}^T_{j, k}\pmb{x}^i_k)}}
\end{equation}
where 
$k$ is the index of features where $k\in[1,...,6]$ corresponds to the 6 local features and $k=7$ corresponds to the global feature, $\pmb{W}_{j,k}$ is the weight vector for identity $j$, and $\pmb{x}_k$ is the extracted feature from each component.

To further improve the performance and speed up the convergence, we apply the batch-hard triplet loss \cite{in_defense_triplet}.
Each mini-batch, consisting of N images, is selected with P identities and K images from each identity.
\begin{equation} \label{triplet_loss}
\begin{split}
L^{1}_{\rm triplet} = \frac{1}{PK}\sum^{P}_{i=1}\sum^{K}_{a=1}[m &+  \max_{p=1...K}\lVert \pmb{x}^{(i)}_a - \pmb{x}^{(i)}_p \rVert_2 \\&-  \min_{\substack{n=1...K \\ j=1...P \\ j \neq i}}\lVert \pmb{x}^{(i)}_a - \pmb{x}^{(j)}_n \rVert_2]_+
\end{split}
\end{equation}
where $\pmb{x}^{(i)}_a$, $\pmb{x}^{(i)}_p$, and $\pmb{x}^{(j)}_n$ are the concatenated and normalized final feature vectors which are extracted from anchor, positive, and negative samples respectively, and $m$ is the margin that restricts the differences between Intra and inter-class distances.

To further ensure the cross-view consistency, we also calculate a triplet loss $L^{2}_{\rm triplet}$ on a 512-dimensional feature vector extracted from the feature map after applying the first attention mask.

By combining all the above losses, our final objective for end-to-end training can be written as minimizing the loss function below:
\begin{equation} \label{combined_loss}
    L_{\rm combined} = L_{\rm ID} + \lambda_1 L^{1}_{\rm triplet} + \lambda_2 L^{2}_{\rm triplet} + \lambda_3 L_{\rm camera}
\end{equation}
where $\lambda_1$, $\lambda_2$ and $\lambda_3$ are used to balance between the classification loss, triplet loss, and camera loss.

\begin{figure}
    \centering
    \begin{subfigure}{0.09\textwidth}
        \includegraphics[width=\textwidth]{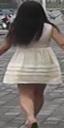}
        \caption{}
    \end{subfigure}
    \begin{subfigure}{0.09\textwidth}
        \includegraphics[width=\textwidth]{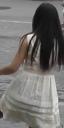}
        \caption{}
    \end{subfigure}
    \begin{subfigure}{0.09\textwidth}
        \includegraphics[width=\textwidth]{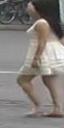}
        \caption{}
    \end{subfigure}
    \begin{subfigure}{0.09\textwidth}
        \includegraphics[width=\textwidth]{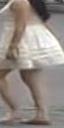}
        \caption{}
    \end{subfigure}
    \begin{subfigure}{0.09\textwidth}
        \includegraphics[width=\textwidth]{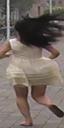}
        \caption{}
    \end{subfigure}
    \caption{An example of imperfect bounding box detection in Market-1501 dataset. (a) is well detected. (b) the bottom part of body has been cropped out. (c) too much background has been included at the bottom. (d) top part is missing. (e) too much background has been included at the top. Imperfect bounding box detection causes misalignment problem which is particularly noxious to strip-based re-ID models.}
    \label{fig:misalignment}
\end{figure}

\subsection{Gaussian Horizontal Data Augmentation}
A major issue that strip-based models cannot circumvent is misalignment. PCB baseline equally slices the last feature map into local strips. Although being focused, the receptive field of each strip actually covers a large fraction of an input image. That is, each local strip can still `see' at least an intact part of the body. Thus, even without explicitly varying feature scales, such as fusing pyramid features or assembling multiple branches with different granularities, the potential of our baseline network to handle misalignment is still theoretically guaranteed.


So the remaining question is how to generate new data mimicking the imperfections of bounding box detection. Some examples of problematic detection that can cause misalignment found in Market-1501 dataset is shown in Figure~\ref{fig:misalignment}. Since the feature cutting is along the vertical direction and global pooling is applied on each strip, the baseline model is more sensitive to the vertical misalignment than the horizontal counterpart. Thus a commonly used random cropping/padding data augmentation is sub-optimal in this case. Instead, we propose a horizontal data augmentation strategy. To be specific, we only randomly crop/pad the top or bottom of the input bounding boxes, by a fraction of the absolute value of a float number drawn from a Gaussian distribution with mean 0 and standard deviation $\sigma$. That is, we assume the level of inaccurate detection follows a form of Gaussian distribution. In all our experiments, the standard deviation $\sigma$ is set to 0.05. This fraction is further clipped at 0.15 to prevent generating outliers. Cropping is adopted when the random number is negative, otherwise, padding is applied. Only with a probability of 0.4, the input images will be augmented in the above way.

\section{\uppercase{Experiments and Results}}
\subsection{Datasets and Evaluation Metrics}
We conduct extensive tests to validate our proposed method on three publicly available person ReID datasets.

\noindent\textbf{Market-1501:} This dataset \cite{market1501} consists of 32,668 images of 1,501 labeled persons captured from 6 cameras. The dataset is split up into a training set which contains 12,936 images of 751 identities, and test set with 3,368 query images and 19,732 gallery images of 750 identities. 

\noindent\textbf{DukeMTMC-reID:}
This dataset is a subset of DukeMTMC \cite{DukeMTMC} which contains 36,411 images of 1,812 persons captured by 8 cameras. 16,522 images of 702 identities were selected as training samples, and the remaining 702 identities are in the testing set consisting of 2,228 query images and  17,661 gallery images.  

\noindent\textbf{CUHK03:} CUHK03 \cite{CUHK03} consists of 14096 images from 1467 identities. The whole dataset is captured by six cameras and each identity is observed by at least two disjoint cameras. In this paper, we follow the new protocol \cite{reranking} which divides the CUHK03 dataset into a training/testing set similar to Market-1501. 

\noindent\textbf{Evaluation Metrics:} To evaluate each component of our proposed model and also compare the performance with existing state-of-the-art methods, we adopt Cumulative Matching Characteristic(CMC) \cite{CMC} at rank-1 and Mean Average Precision(mAP) in all our experiments. Note that all the experiments are conducted in a single-query setting without applying re-ranking \cite{reranking}.

\subsection{Implementation Details}

\noindent\textbf{Data Pre-processing:} During training, the input images will be re-sized to a resolution of $384\times 128$ to better capture detailed information. We deploy random horizontal flipping and random erasing \cite{zhong17} for data augmentation. Note that our complete framework contains a horizontal data augmentation which will be deployed before image re-sizing.

\noindent\textbf{Loss Hyper-parameters:} In all our experiments, we set the parameter of label smoothing softmax loss $\epsilon=0.1$. Because our classification loss is the addition of global classification loss and local classification loss, so we give weight to the triplet loss. The parameters for the combined loss are set to $\lambda_1=5$, $\lambda_2=5$ and $\lambda_3=1$. Here we set $P=24$ and $K=4$ in triplet loss to train our proposed model.

\noindent\textbf{Optimization:} We use SGD with momentum 0.9 to optimize our model. The weight decay factor is set to 0.0005. To let the components that haven't been pre-trained get up to speed, we set the initial learning rate of attention modules, feature extractors, and classifiers to 0.1, while we set the initial learning rate of the backbone network to 0.01. The learning rate will be dropped by half at epochs 150, 180, 210, 240, 270, 300, 330, 360, and we let the training run for 450 epochs in total.

\subsection{Ablation Study}
We further perform comprehensive ablation studies with each component of our proposed model on Market-1501 datasets.

\begin{table}[h]
\centering
    \caption{Evaluating each component in our proposed method.}
    \begin{tabular}{l|r|r}
        \hline
        \multicolumn{1}{l|}{Dataset} & \multicolumn{2}{c}{Market-1501}\\
        \cline{1-3}\multicolumn{1}{l|}{Metric(\%)} & \multicolumn{1}{l|}{\textit{rank 1}} & \multicolumn{1}{l}{\textit{mAP}} \\
        \hline
        Baseline &93.2&82.2\\
        Base+MRFA &93.8&83.2\\
          \hspace{0.5cm}- features before $\otimes$+CAM &93.3 & 82.8\\
          \hspace{0.5cm}- features after  $\otimes$+CAM &93.3 & 83.1\\
        Base+MRFA+CAM &94.3&83.9\\
        Base+MRFA+CAM+TL &95.2&87.5\\
        Base+MRFA+CAM+TL+HDA &\textbf{95.5}&\textbf{88.1}\\
    \hline
    \end{tabular}
    \label{tab:ablation_study}
\end{table}

\noindent\textbf{Benefit of Attention Modules:}
We first evaluate the effect of our proposed multi-receptive field attention (MRFA) module by comparing it with the baseline network. 
The results are shown in table~\ref{tab:ablation_study}. We observe an improvement of $0.6\%/1.0\%$ \textit{rank 1/mAP} on Market-1501. Notice that MRFA is only added to the last two stages of the ResNet50 baseline. We observe little improvements when adding MRFA to the front stages of the backbone network. Considering the cost of a more complicated network, we decide to only add MRFA on the last two stages. 


\begin{table*}[t]
  \centering
  \caption{Comparison with the state-of-the-arts on Market-1501 and DukeMTMC-ReID datasets. The best results are in bold, while the numbers with underlines denote the second best.}
    \begin{tabular}{l|r|r|r|r}
    \hline
    \multicolumn{1}{c|} {\multirow{2}[4]{*}{Model}} & \multicolumn{2}{c|}{Market1501} &  \multicolumn{2} {c}{DukeMTMC-reID} \\
    \cline{2-5}          & \multicolumn{1}{l|}{\textit{rank 1}} & \multicolumn{1}{l|}{\textit{mAP}} & \multicolumn{1}{l|}{\textit{rank 1}} & \multicolumn{1}{r}{\textit{mAP}}  \\
    \hline
    SVDNet\cite{svdnet} &    82.3   &  62.1     &   76.7    &  56.8       \\
    PAN\cite{pan}   &   82.8    & 63.4      &    71.6  &  51.5     \\
    MultiScale\cite{multiscale} &     88.9  &   73.1  &    79.2   & 60.6      \\
    \hline
    MLFN \cite{mlfn} &   90.0    & 74.3    &   81.0   &     62.8   \\
    HA-CNN\cite{hacnn} &    91.2   &    75.7   &    80.5   &    63.8     \\
    Mancs\cite{mancs} &    93.1   &    82.3   &    84.9   &    71.8     \\
    Attention-Driven\cite{YANG2019143} &    94.9   &    86.4   &    86.0   &    74.5     \\
    \hline
    PCB+RPP\cite{pcb} &   93.8    & 81.6      &     83.3  & 69.2      \\
    HPM  \cite{horizontal} &   94.2    &    82.7   &    86.6   &  74.3      \\
    MGN \cite{mgn}  &   \textbf{95.7}    &  \underline{86.9}     &     \underline{88.7}  &    \underline{78.4}     \\
    \hline
    VMRFANet(Ours)  & \underline{95.5} & \textbf{88.1} & \textbf{88.9} & \textbf{80.0} \\
    \hline
    \end{tabular}
  \label{tab:comparison}
\end{table*}

\begin{table*}[t]
  \centering
  \caption{Comparison of results on CUHK03-labeled (CUHK03-L) and CUHK03-detected (CUHK03-D) with new protocol \cite{reranking}. The best results are in bold, while the numbers with underlines denote the second best.}
    \begin{tabular}{l|r|r|r|r}
    \hline
    \multicolumn{1}{c|}{\multirow{2}[4]{*}{Model}} & \multicolumn{2}{c|}{CUHK03-L} & \multicolumn{2}{c}{CUHK03-D} \\
\cline{2-5}          & \multicolumn{1}{l|}{\textit{rank 1}} & \multicolumn{1}{l|}{\textit{mAP}} & \multicolumn{1}{l|}{\textit{rank 1}} & \multicolumn{1}{l}{\textit{mAP}} \\
    \hline
    SVDNet\cite{svdnet} & 40.9  & 37.8  & 41.5  & 37.3 \\
    MLFN\cite{mlfn}  & 54.7  & 49.2  & 52.8  & 47.8 \\
    HA-CNN\cite{hacnn} & 44.4  & 41.0    & 41.7  & 38.6 \\
    PCB+RPP\cite{pcb} & --      &--      & 63.7  & 57.5 \\
    MGN\cite{mgn}   & 68.0    & 67.4  & 68.0    & 66.0 \\
    \hline
    MRFANet (Ours) & \textbf{81.1}    & \textbf{78.8}  & \textbf{78.9}  & \textbf{75.3} \\
    \hline
    \end{tabular}
  \label{tab:CUHK03}
\end{table*}

\noindent\textbf{Effectiveness of View-specific Learning:} We compare the performance of our proposed model with and without adding the camera ID classification loss to the MRFA modules (see first and the last row of table~\ref{tab:ablation_study}). We see 0.5\%/0.7\% gain at \textit{rank 1/mAP} on Market1501 with view specific learning on attention mechanism.

To further show the necessity for adding camera loss on attention mechanism and the primary cause of the performance gain is not simply because of introducing a harder objective, we conduct experiment moving two camera losses from attention mechanism to features of corresponding stages (stage 3 and stage 4) of the backbone network. We experiment two settings, one is to add camera loss before $\otimes$ operation with attention and another is to add camera loss after $\otimes$ operation. In both setting (see fourth and fifth rows in table~\ref{tab:ablation_study}) , we see degradation on \textit{rank 1} and \textit{mAP}. It demonstrated that adding camera loss directly on the backbone network is not helpful. It likely affects the camera-invariant features extracted by the backbone network.


\noindent\textbf{Benefit of Combined Objective Training with Triplet and Softmax Loss:}
Our network is trained by minimizing both triplet loss and softmax loss jointly. We evaluated its performance comparing to our baseline+MRFA+CAM setting. We found that the combination of losses not only brings significant improvements ($+0.9\%/+3.6\%$ \textit{rank 1/mAP} on Market-1501) on the performance but also speeds up the convergence. Notably, the triplet loss is essential since it serves as the cross-view consistency regularization term in the view-specific learning mechanism. 

\noindent\textbf{Impact of Horizontal Data Augmentation on Strip-based Re-ID Model:}
Finally, we add horizontal data augmentation to the network Baseline+MRFA+CAM and get our final view-specific multi-receptive field attention network (VMRFANet: Baseline+MRFA+CAM+HDA).  We do the comparisons of the models with and without horizontal data augmentation. The performance gain ($+0.3\%/+0.6\%$ \textit{rank 1/mAP} on Market-1501 dataset) proves the effectiveness of the data augmentation strategy against misalignment.


\subsection{Comparison with State-of-the-art}
We evaluate our proposed model against current state-of-the-arts methods on three large benchmarks. The comparisons on Market-1501 and DukeMTMC-reID are summarized in Table~\ref{tab:comparison}, while the results on CUHK03 is shown in Table~\ref{tab:CUHK03}.

\noindent\textbf{Results on Market-1501:} Our method achieves the best result on \textit{mAP} metric, and the second best on \textit{rank 1}. It outperforms all other approaches except a strip-based method MGN \cite{mgn} on \textit{rank 1} metric. However, MGN incorporates three independent branches after stage 3 of the ResNet50 backbone to extract features with multi-granularity. Moreover, the difference is only marginal, and our method has achieved this competitive result using a much smaller network. Remarkably, on this dataset whose bounding boxes are automatically detected, the Gaussian horizontal data augmentation strategy greatly improves the robustness of the model. 


\noindent\textbf{Results on DukeMTMC-reID:} Our method achieves the best results on this dataset at both metrics. Notably, PCB \cite{pcb} is a strip-based model that serves as the starting point of our approach. We surpassed it by $+10.8\%$ on \textit{mAP} and $+5.6\%$ on \textit{rank 1}. MGN gets the second best results among all compared methods on this dataset. On the other hand, our model outperforms the listed attention-based models by a large margin.

\noindent\textbf{Results on CUHK03:}
To evaluate our proposed method on CUHK03, we follow the new protocol \cite{reranking}. However, since only a relative label (with binary values 1 and 2) is used for identifying which camera that an image is coming from, we found it hard to extract the exact camera IDs from CUHK03. Thus we only test our model without enabling the view-specific learning on this dataset. In table~\ref{tab:CUHK03}, we show the results of our proposed method on CUHK03. Remarkably, although the MRFA module is not guided by camera ID, our model still outperforms all other methods by a large margin.

\section{\uppercase{Conclusion}}
In this work, we introduce
a novel multi-receptive field attention module which brings a considerable performance boost to a strip-based person re-ID network. Besides, we propose a horizontal data augmentation strategy which is shown to be particularly helpful against misalignment issues. Combined with the idea of injecting view information through the attention module, our proposed model achieves superior performance comparing to current state-of-the-art on three widely used person re-identification benchmark datasets.

\bibliographystyle{apalike}
{\small
\bibliography{example}}

\end{document}